%% file: main.tex
\documentclass[letterpaper, 10 pt, conference]{ieeeconf}  

\usepackage{times}
\usepackage{amsmath}
\usepackage[pdftex]{graphicx}
\usepackage{caption}
\usepackage{subcaption}
\usepackage{amsmath,amssymb,amsopn,amstext,amsfonts}
\usepackage{cancel}
\usepackage[space]{cite}
\usepackage{pdfsync}
\usepackage{balance}
\usepackage{color}
\usepackage{algorithm}
\usepackage{algorithmic}
\usepackage{url}
\usepackage{booktabs}
\usepackage{threeparttable}
\usepackage[linkcolor=black,citecolor=black,urlcolor=black,colorlinks=true]{hyperref}
\usepackage{multirow}
\usepackage[outdir=./epstopdf/]{epstopdf}
\usepackage{graphicx}
\usepackage{array}
\usepackage{verbatim}
\usepackage{makecell}
\usepackage{cleveref}
\usepackage{adjustbox}
\usepackage{array}
\usepackage{booktabs}
\usepackage{multirow}
\usepackage{pifont}
\usepackage{color}
\usepackage{balance}
\usepackage{stfloats}

\IEEEoverridecommandlockouts                              

\DeclareGraphicsExtensions{.pdf,.png,.jpg,.eps,.PNG}
\title{\LARGE \bf
Enhancing Exploratory Capability of Visual Navigation Using Uncertainty of Implicit Scene Representation
}

\author{Yichen Wang, Qiming Liu, Zhe Liu, and Hesheng Wang*
\thanks{This work was supported in part by the Natural Science Foundation of China under Grant 62225309, 62303307, 62073222, U21A20480 and 62361166632.}
\thanks{Y. Wang and Q. Liu are with the Department of Automation, Shanghai Jiao Tong University, Shanghai 200240, China. Z. Liu is with MoE Key Lab of Artificial Intelligence, AI Institute, Shanghai Jiao Tong University, Shanghai 200240, China. H. Wang is with the Department of Automation, Shanghai Jiao Tong University, Shanghai 200240, China.}
\thanks{*Corresponding author: Hesheng Wang ({e-mail: \tt\small wanghesheng@
sjtu.edu.cn}).}
}

\begin{document}

\maketitle
\thispagestyle{empty}
\pagestyle{empty}


\input{sec/Abstract}


\input{sec/Introduction}

\input{sec/Related_Works}

\input{sec/Methodology}

\input{sec/Experiment}

\input{sec/Conclusion}


{
    \small
    \bibliography{main}
}

\balance

\end{document}

%% file: sec/Abstract.tex
\begin{abstract}
In the context of visual navigation in unknown scenes, both ``exploration'' and ``exploitation'' are equally crucial. Robots must first establish environmental cognition through exploration and then utilize the cognitive information to accomplish target searches. However, most existing methods for image-goal navigation prioritize target search over the generation of exploratory behavior. To address this, we propose the Navigation with Uncertainty-driven Exploration (NUE) pipeline, which uses an implicit and compact scene representation, NeRF, as a cognitive structure. We estimate the uncertainty of NeRF and augment the exploratory ability by the uncertainty to in turn facilitate the construction of implicit representation. Simultaneously, we extract memory information from NeRF to enhance the robot's reasoning ability for determining the location of the target. Ultimately, we seamlessly combine the two generated abilities to produce navigational actions. Our pipeline is end-to-end, with the environmental cognitive structure being constructed online. Extensive experimental results on image-goal navigation demonstrate the capability of our pipeline to enhance exploratory behaviors, while also enabling a natural transition from the exploration to exploitation phase. This enables our model to outperform existing memory-based cognitive navigation structures in terms of navigation performance. Project page: \href{https://github.com/IRMVLab/NUE-NeRF-nav}{https://github.com/IRMVLab/NUE-NeRF-nav}
\end{abstract}

%% file: sec/Introduction.tex
\section{Introduction}
\label{sec:intro}

\begin{figure}[t] 
	\centering 
	\includegraphics[width=\linewidth]{"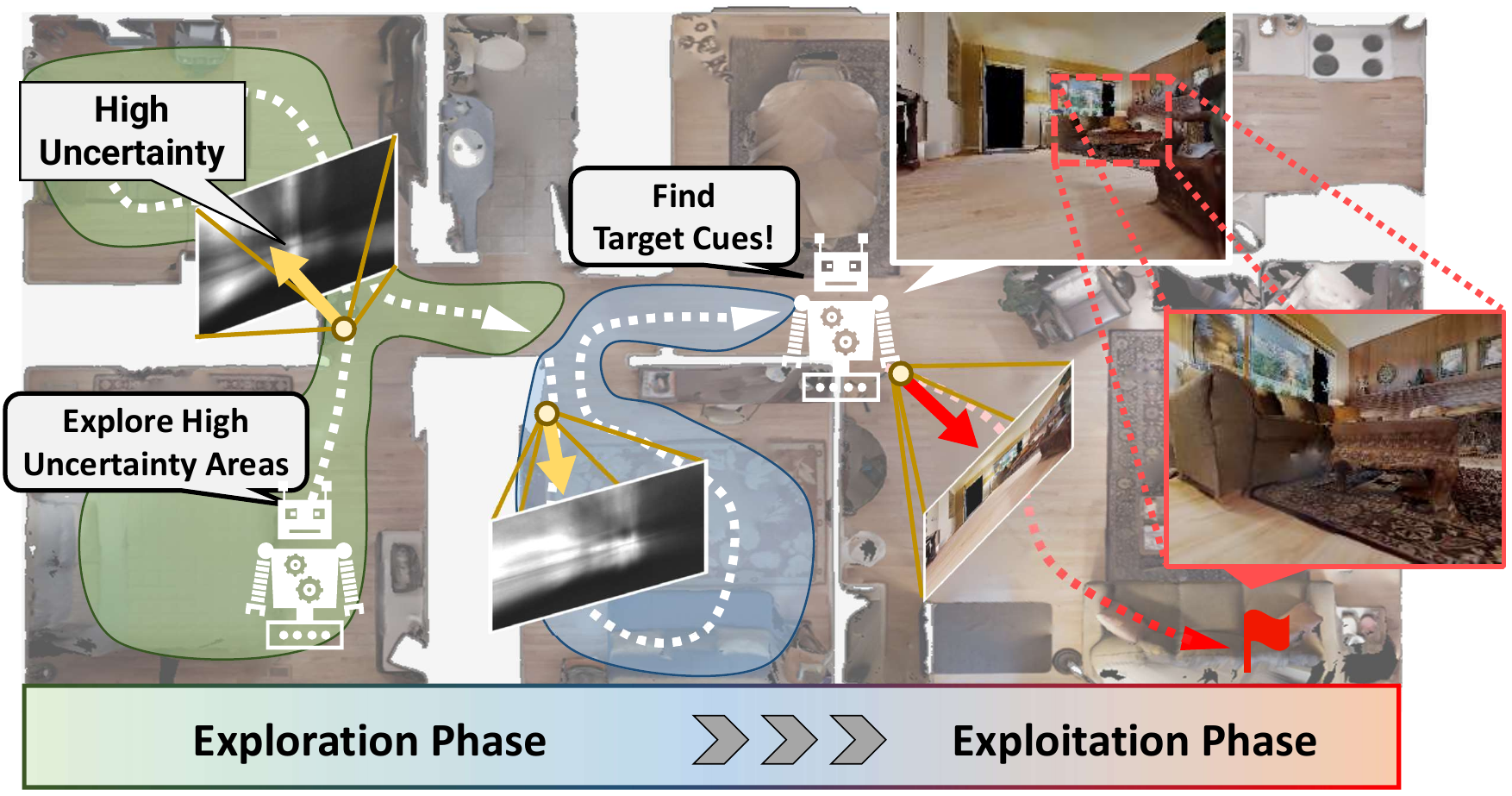"} 
	\caption{\textbf{Navigation of robots in an unknown environment.} In visual navigation of robots in unknown environments, the process involves two phases: exploration and exploitation. Initially, the robot explores based on uncertainty to refine its cognitive structure, transitioning to navigation toward detected target-related cues in the environment.} 
    \label{fig_1}
\vspace{-\baselineskip}
\end{figure}

When searching for a target in an unfamiliar environment, our subconscious instinctively prioritizes exploring to establish cognitive. Upon locating the target, we shift to exploitation, navigating towards it. When a robot engages in visual navigation tasks in an unknown environment, as shown in Fig.~\ref{fig_1}, it also benefits from both exploratory and exploitative thinking. However, existing cognitive navigation frameworks primarily focus on the robot's performance in the exploitation phase, neglecting the design of its exploratory behavior. We intend to use implicit scene representation as the memory structure of our navigation pipeline, specifically emphasizing the enhancement of the robot's exploratory capability. This enhancement enables the robot to rapidly establish environmental awareness, discover target-related cues, and transition into the exploitation phase.

\begin{figure}[t] 
	\centering 
	\includegraphics[width=\linewidth]{"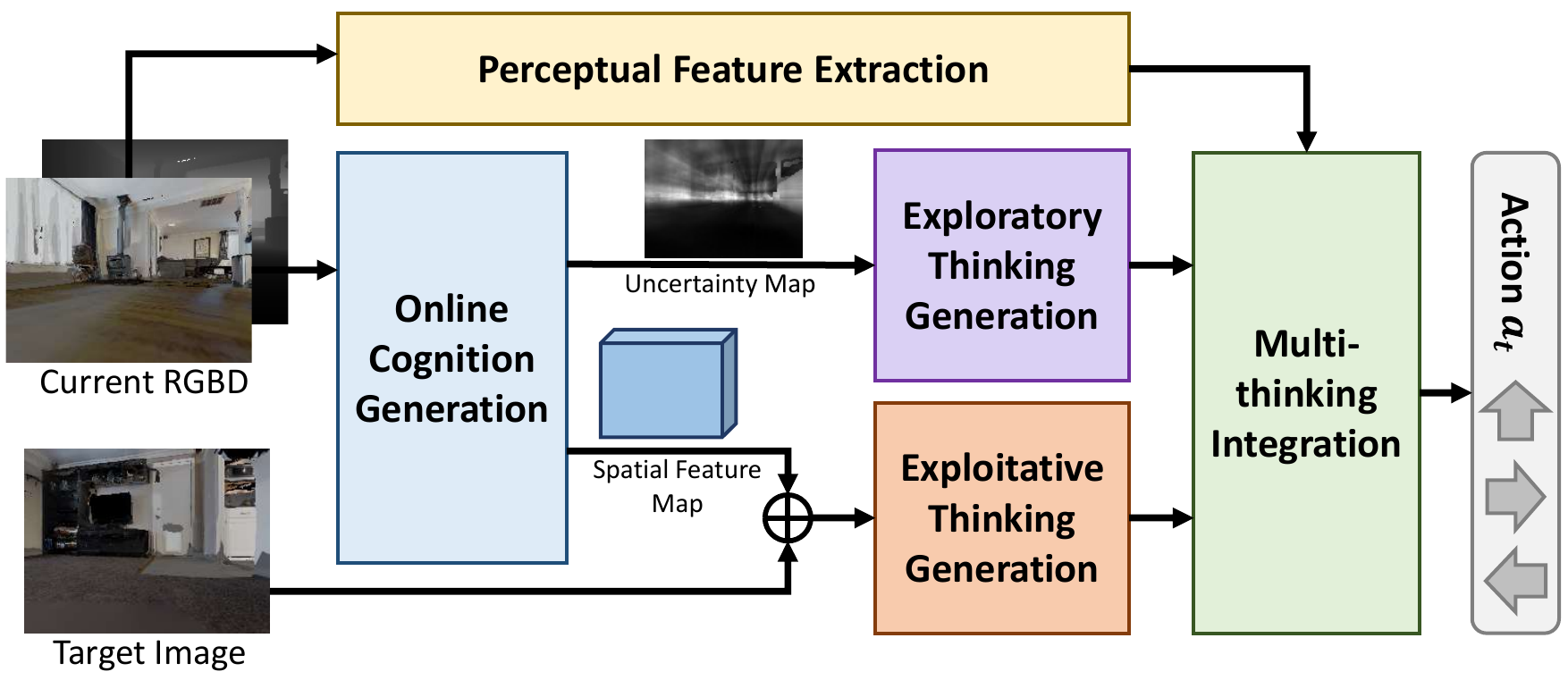"} 
	\caption{\textbf{The overall architecture of NUE.} Firstly, real-time image input is used for online cognitive generation and perceptual feature extraction. Secondly, cognitive information is extracted to generate exploratory thinking and exploitative thinking. Eventually, multiple thinking is integrated, and navigational actions are generated.}
    \label{fig_2}
\vspace{-\baselineskip}
\end{figure}

In this paper, we adopt the Neural Radiance Fields (NeRF)~\cite{ref_1} as our memory structure. In recent years, NeRF has shown excellent performance in reconstructing 3D scenes and synthesizing novel views~\cite{ref_6}. Due to its compact implicit scene representation capability and network-based structure, NeRF has shown great potential in downstream navigation tasks that rely on long-term information representation. However, since our optimization of NeRF is performed online, its high signal-to-noise output early in navigation struggles to provide valuable environmental memory information. To overcome this challenge, we use an estimate of NeRF's uncertainty and leverage it to foster exploratory thinking within the robot, considering that exploration can speed up the establishment of implicit representations.

We propose the Navigation with Uncertainty-driven Exploration (NUE) pipeline, as illustrated in Fig.~\ref{fig_2}, which is end-to-end and fully differentiable. Firstly, we conduct online training on NeRF to generate cognition of the scenes. To enhance the exploration capability of the robot and reduce the noise of NeRF in the initial navigation phase, we include an estimate of NeRF's uncertainty. Secondly, we extract distinct components, namely uncertainty and spatial information, from NeRF. These components are subsequently compressed into feature representations. The uncertainty feature is leveraged to augment exploration capabilities, while the spatial feature is harnessed to optimize exploitation performance. Finally, we adaptively fuse the two features and output navigation actions through an action generator. Our main innovations can be summarized as follows:

\begin{itemize}
    \item We propose NUE, an end-to-end visuomotor navigation pipeline integrating NeRF as a cognitive structure. By leveraging the compact scene representation capabilities of NeRF, we extend its application from the perception domain to the control domain.
    \item We utilize the estimation of NeRF's uncertainty to enable the robot to exhibit exploratory behavior. Additionally, our model successfully balances exploratory and exploitative thinking, achieving seamless integration between the exploration and exploitation stages.
    \item Experimental results demonstrate that NUE significantly improves navigation performance compared to existing cognitive memory structures. Interpretability experiments validate that NUE effectively generates and balances exploration and exploitation behaviors.
\end{itemize}

%% file: sec/Related_Works.tex
\section{Related Works}
\label{sec:Related works}

\subsection{Neural Radiance Fields in Perception}

Neural Radiance Fields (NeRF)~\cite{ref_1} is an innovative approach for synthesizing views. It combines multi-layer perceptrons (MLPs) with volume rendering techniques to generate new views. What makes NeRF impressive is its ability to achieve high-quality scene representation with a compact and concise structure. Currently, many efforts are focused on improving the training and inference efficiency of NeRF and addressing the issue of sparse view sensitivity.

Due to the computationally intensive of NeRF, some methods focus on improving the structure of NeRF~\cite{ref_32} or exploring the use of additional depth supervision~\cite{ref_36} to accelerate the inference process. In addition, the high-quality scene reconstruction in NeRF relies on a large number of densely sampled visual observations. However, obtaining densely sampled views can be challenging in many tasks. Therefore, some works~\cite{ref_39} have focused on improving NeRF's sensitivity to sparse views. This is crucial for navigation tasks as well; since robots often have limited and unevenly distributed local observations during navigation.

In addition, several studies have applied NeRF to SLAM (simultaneous localization and mapping) systems. iNeRF~\cite{ref_40} first models pose estimation as the inverse process of NeRF inference. Based on this, some studies have improved the sampling method~\cite{ref_43}, further optimizing the performance and efficiency of pose estimation. While these studies have achieved impressive results in the perception domain, they have not yet extended the powerful characterization capabilities of NeRF to the control domain.

\subsection{Neural Radiance Fields in Robotics}
In the field of robotics, NeRF is mainly applied to robotic arm control~\cite{ref_44,ref_46} and navigation~\cite{ref_47,ref_48,ref_51,ref_52}. In the field of navigation, certain studies strive to establish the transformational relationship between NeRF and the geometric representation of the occupancy space. This enables precise estimation of the scene's geometric structure, and, by utilizing planning methods, facilitates the generation of collision-free and smooth trajectories~\cite{ref_47,ref_48}. Although these works have made great progress in navigation, they still rely on using NeRF as an auxiliary tool for trajectory planning and have not fully integrated the implicit structure of NeRF with neural controllers.

Some recent studies~\cite{ref_51,ref_52} have attempted to connect the implicit representation directly to neural controllers and output navigation actions. Additionally, researchers are focusing on the problem of active representation learning, which explores how to use navigation robots to better construct NeRF representations through exploration in the environment~\cite{ref_53,ref_57}. While these studies have achieved remarkable results, they often lack a natural integration of the exploration and exploitation phases. Our focus is on enabling the robots to learn a multi-modal thinking approach, allowing for the generation of a navigation strategy that closely aligns with human thinking patterns.

\begin{figure*}[t] 
	\centering 
	\includegraphics[width=0.95\linewidth]{"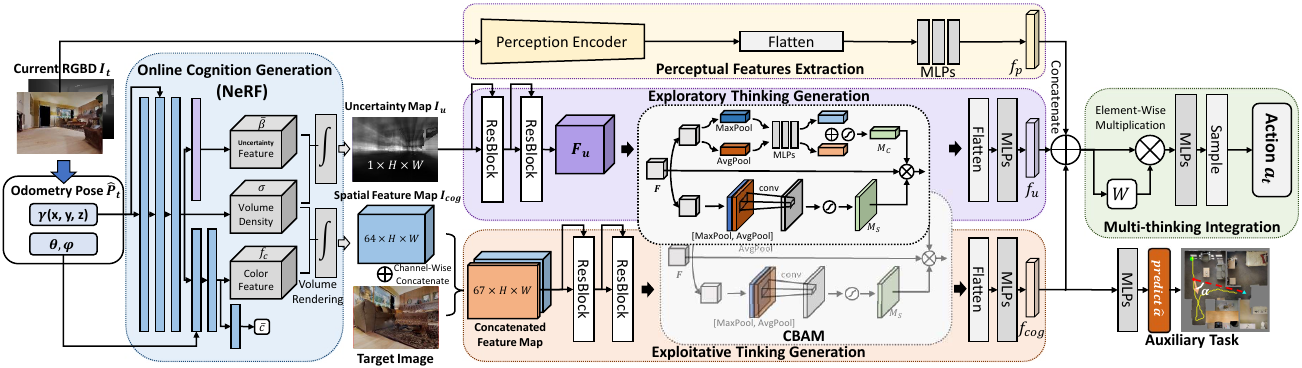"} 
	\caption{\textbf{The network structure of NUE.} The overall framework first extracts real-time perceptual features and generates cognitive signals in NeRF. Subsequently, we compress the uncertainty map to generate uncertainty features and concatenate the spatial feature map with the target image in the channel dimension for spatial feature extraction. Finally, the features are concatenated and fed into an adaptive neural controller to generate the final navigation actions.}
    \label{fig_3}
\vspace{-\baselineskip}
\end{figure*}

%% file: sec/Methodology.tex
\section{Methodology}
\label{sec:Methodology}

\subsection{Overall System Architecture}
We aim to enhance the spatial cognition capability of robots to achieve better exploratory behavior. To achieve this goal, we introduce NUE, as shown in Fig.~\ref{fig_3}, which consists of three key processes. The first part involves online cognition generation, where the robot stores real-time perceptual information of the environment in NeRF, providing the robot with spatial cognition. The second part is online cognitive extraction, which utilizes resnet and CBAM~\cite{ref_58} to extract features from the uncertainty and spatial information produced by NeRF, and generate corresponding exploration and exploitation strategies. The third part encompasses multi-thinking integration, extracting perceptual features from real-time image input and then fusing perceptual features, uncertainty features, and spatial features to generate navigational actions through a neural controller.

\subsection{Online Cognition Generation}
The spatial cognition of robots is generated online, independent of any prior knowledge of the scenes. However, due to the online generation nature, the model's cognitive awareness has significant uncertainty in the early stages of the navigation process, which is not conducive to direct information utilization. Therefore, we enhance exploration by estimating the uncertainty of the cognitive structure.

We use NeRF to generate the spatial cognition of the robot. To incorporate uncertainty estimation into the model, we combine previous research~\cite{ref_39} and model the emitted radiance values at each position in space as a Gaussian distribution, parameterized by mean $\bar{c}$ and variance $\bar{\beta}^2$. The formulation is as follows:
\begin{equation}\label{e1}
\left [ \sigma , f , \bar{\beta}^2(\textrm{r}(t)) \right ] = \textrm{MLP}_{\theta _{1}, \theta _{3}}(\gamma _{x}(\textrm{r}(t))),
\end{equation}
\begin{equation}\label{e2}
\bar{c}(\textrm{r}(t))  = \textrm{MLP}_{\theta _{2} }(\it{f}, \gamma _{d}(\textrm{r}(t))),
\end{equation}
\noindent where $\gamma _{x, d}(\cdot)$ is the position encoding functions, $f$ represents the intermediate features, and $\textrm{r}(t)=o+td$ represents the entire ray, with $o$ as the origin and $d$ as the direction of the ray from the camera center to the sampled point.

Due to the incorporation of uncertainty, we adopt the loss function used in ActiveNeRF~\cite{ref_39} to optimize the NeRF network. The main loss function is as follows:
\begin{equation}\label{e5}
\mathcal{L}^{u}_{i} = \frac{\left \| C(\textrm{r}_i)-\bar{C} (\textrm{r}_i) \right \|_{2}^{2} }{2\bar{\beta}^{2}(\textrm{r}_i)}+\frac{\log{\bar{\beta}^2(\textrm{r}_i)}}{2},
\end{equation}
\noindent where $\textrm{r}_i$ is sampled ray. $\bar{C}(\textrm{r}_i)$ and $\bar{\beta}^2(\textrm{r}_i)$ denote the mean and variance of rendering colors through sampled ray $\textrm{r}_i$, respectively. $C(\textrm{r}_i)$ denotes the ground truth.


\subsection{Online Cognition Extraction}
In this section, we aim to imbue the robot with an integration of both exploratory and exploitative cognition. This involves prioritizing unexplored areas before observing the target, collecting target-related cues, and showing a preference for the target once it is observed. We achieve this by extracting uncertainty features for exploration and spatial features from the cognitive structure for exploitation.

\subsubsection{Exploratory Thinking Generation}
To generate exploratory thinking, we extract the uncertainty features from NeRF. We render the uncertainty $\bar{\beta}^{2}(\textrm{r}(t_i))$ outputted by NeRF as an uncertainty map $I_u\in \mathbb{R} ^{1 \times W\times H}$, the rendering of uncertainty can be described as follows:
\begin{equation}\label{e6}
I_{u}(\textrm{r}) = \sum_{i = 1}^{N_s}\alpha_i \bar{\beta}^{2}(\textrm{r}(t_i)), 
\end{equation}
\begin{equation}\label{e7}
\alpha _i = \textrm{exp}( -\sum_{j=1}^{i-1}\sigma _j \delta _j  ) \left ( 1-\textrm{exp}\left ( \sigma _i \delta _i \right )  \right ),
\end{equation}
\noindent where $\delta _j=t_{i+1}-t_{i}$ represents the distance between neighboring sampling points, and $N_s$ is the number of sampling points on each ray.

Uncertainty map $I_u$ can reflect the familiarity of NeRF with different regions of the scene in the current perspective. As shown in Fig.~\ref{fig_3}, we first use two cascaded residual structures to extract features from the uncertainty map $I_u$. Through the resnet network, we compress the input uncertainty map $I_u$ into a feature map $F_u\in \mathbb{R} ^{16 \times W/16 \times H/16}$.

After compressing $I_u$ to $F_u$, we employ channel attention and spatial attention to adaptively optimize the feature map.

~\\
\noindent\textbf{Channel Attention.} Channel attention compresses the input feature maps into $F_{avg}^{c}$ and $F_{max}^{c}$ by aggregating the spatial information using average pooling and maximum pooling in the spatial dimension. The process is as follows:

\begin{equation}\label{e8}
M_C(F) = \textrm{Sigmod} \left ( \textrm{MLP}\left ( F_{avg}^{c} \right ) + \textrm{MLP}\left ( F_{max}^{c} \right )\right ), 
\end{equation}

\noindent\textbf{Spatial Attention.} Spatial attention complements channel attention by exploiting average pooling and maximum pooling operations on the channel dimension to compress the feature graph into $F_{avg}^{c}$ and $F_{max}^{c}$. The specific process is as follows:

\begin{equation}\label{e9}
M_S(F) = \textrm{Sigmod} \left ( \textrm{Conv}\left ( \left [  F_{avg}^{s}, F_{avg}^{s}\right ]  \right ) \right ), 
\end{equation}

We multiply the two attention maps $M_C$, $M_S$ with the original feature map $F_u$. Finally, the feature map is further compressed into a perceptual feature vector $f_u$ of length 64 through an MLP. This process can be represented as follows:
\begin{equation}\label{e10}
f_u = \textrm{MLP}(\textrm{Flatten}(M_C \otimes M_S \otimes F_u)),
\end{equation}
\noindent where $\otimes$ denotes element-wise multiplication

\subsubsection{Exploitative Thinking Generation}

To generate exploitative thinking, we extract the spatial features from NeRF. We first generate the volume density $F_{\sigma}\in \mathbb{R} ^{1 \times W\times H\times N_s}$ and the feature map $F_c\in \mathbb{R} ^{64 \times W\times H\times N_s}$ from the intermediate layer through NeRF. After obtaining the voxel density and color features for all sampling points, we perform voxel rendering on the color features to generate a compressed spatial feature map $I_{cog}\in \mathbb{R} ^{64 \times W\times H}$. The specific procedure is described as follows:
\begin{equation}\label{e11}
f_c  = \textrm{MLP}_{\theta _{2} }(\it{f}, \gamma _{d}(\textrm{r}(t))),
\end{equation}
\begin{equation}\label{e12}
I_{cog}(\textrm{r})=\sum_{i=1}^{N_s}\alpha_i {f_c}(\textrm{r}(t_i)),
\end{equation}
\noindent where $\textrm{MLP}_{\theta _{2}}$ is the same as in equation \ref{e2}, $N_s$, $\alpha_i$ and $\textrm{r}(t_i)$ is the same as in equation \ref{e6}.

The spatial feature map $I_{cog}\in \mathbb{R} ^{64 \times W\times H}$ contains the structural information of the scene from the current viewpoint. We concatenate the target RGB image with $I_{cog}$ along the channel dimension, and the concatenated feature map undergoes two residual blocks for additional feature extraction. Subsequently, the compressed features are passed through CBAM structure to eliminate noise and amplify navigation-related information. Finally, the cognitive information is compressed into a 64-length feature vector $f_{cog}$ using an MLP. This step associates the spatial cognitive ability of the robot with target navigation, enhancing the robot's navigation capabilities during the exploitation phase.

\noindent \textbf{Auxiliary Task.} We aggregate cognitive information through an implicit process, ensuring an end-to-end characteristic of the network. To improve network interpretability, we introduce an auxiliary task predicting the angle between the robot's current orientation and the target direction. The operation procedure of the auxiliary task is to input the feature vector $f_{cog}$ into the two-layer MLP to obtain the predicted value $\hat{\alpha}$ for the angle of the pinch, and compare $\hat{\alpha}$ and the true value $\alpha$ using the $L1$ loss function, and optimize it by minimizing the loss.

\begin{table*}[t]
\centering
\caption{\textbf{Image-goal navigation results.} The data in bold in each column represents the optimal data for that column. The bottommost row is the test results of our full model, NUE.}
\begin{adjustbox}{width=\textwidth}
\begin{tabular}{l|>{\centering\arraybackslash}p{1cm} >{\centering\arraybackslash}p{0.6cm} >{\centering\arraybackslash}p{1cm}|ccc|ccc|ccc|ccc}
\hline \hline
\multirow{2}{*}{}                               & \multicolumn{3}{c|}{\multirow{2}{*}{}} & \multicolumn{3}{c|}{\textbf{Easy}} & \multicolumn{3}{c|}{\textbf{Medium}} & \multicolumn{3}{c|}{\textbf{Hard}} & \multicolumn{3}{c}{\textbf{Total}} \\ \hline
\multirow{2}{*}{}                               & \multicolumn{3}{c|}{}                  & \textbf{SR}$\uparrow$     & \textbf{SPL}$\uparrow$    & \textbf{DTS}$\downarrow$    & \textbf{SR}$\uparrow$     & \textbf{SPL}$\uparrow$     & \textbf{DTS}$\downarrow$     & \textbf{SR}$\uparrow$     & \textbf{SPL}$\uparrow$    & \textbf{DTS}$\downarrow$    & \textbf{SR}$\uparrow$     & \textbf{SPL}$\uparrow$     & \textbf{DTS}$\downarrow$    \\ \hline
\multicolumn{1}{l|}{\multirow{7}{*}{\rotatebox{90}{\textbf{Baselines}}}} & \multicolumn{3}{c|}{\textbf{FR}~\cite{ref_61}}                & 22.37\%       & 0.213       & 1.635       & 8.96\%       & 0.085        & 4.242        & 8.18\%       & 0.075       & 2.976       & 12.54\%       & 0.012        & 3.049       \\
\multicolumn{1}{l|}{}                           & \multicolumn{3}{c|}{\textbf{Nav-A3C}~\cite{ref_16}}           & 44.83\%       & 0.318       & 1.327       & 37.72\%       & 0.320        & 2.138        & 23.22\%       & 0.200       & 3.245       & 33.90\%       & 0.272        & 2.355       \\
\multicolumn{1}{l|}{}                           & \multicolumn{3}{c|}{\textbf{MSM}~\cite{ref_62}}               & 47.32\%       & 0.260       & \textbf{1.097}       & 25.60\%       & 0.114        & 2.220        & 18.08\%       & 0.117       & 3.440       & 30.50\%       & 0.167        & 2.259       \\
\multicolumn{1}{l|}{}                           & \multicolumn{3}{c|}{\textbf{VGM}~\cite{ref_27}}               & 47.13\%       & 0.295       & 1.313       & 42.04\%       & 0.278        & 2.478        & 36.22\%       & 0.272       & 3.702       & 41.37\%       & 0.295        & 2.590       \\
\multicolumn{1}{l|}{}                           & \multicolumn{3}{c|}{\textbf{NRNS}~\cite{ref_64}}              & 43.00\%       & 0.317       & 1.493       & 30.60\%       & 0.209        & 2.106        & 18.61\%       & 0.133       & 3.575       & 30.73\%       & 0.219        & 2.391       \\
\multicolumn{1}{l|}{}                           & \multicolumn{3}{c|}{\textbf{SLING+DDPPO}~\cite{ref_66,ref_67}}       & 40.00\%       & 0.273       & 2.007       & 36.50\%       & 0.272        & 2.136        & 24.00\%       & 0.205       & 3.191       & 33.50\%       & 0.250        & 2.445       \\
\multicolumn{1}{l|}{}                           & \multicolumn{3}{c|}{\textbf{SLING+OVRL}~\cite{ref_66,ref_68}}              & 63.18\%       & \textbf{0.477}       & 1.385       & 55.00\%       & \textbf{0.395}        & 1.749        & 51.75\%       & 0.267       & 3.269       & 56.64\%       & 0.379        & 2.134       \\ \hline \hline
\multicolumn{1}{l|}{} & \textbf{$f_u$}         & \textbf{AT}        & \textbf{CBAM}        & \textbf{SR}$\uparrow$     & \textbf{SPL}$\uparrow$    & \textbf{DTS}$\downarrow$    & \textbf{SR}$\uparrow$     & \textbf{SPL}$\uparrow$     & \textbf{DTS}$\downarrow$     & \textbf{SR}$\uparrow$     & \textbf{SPL}$\uparrow$    & \textbf{DTS}$\downarrow$    & \textbf{SR}$\uparrow$     & \textbf{SPL}$\uparrow$     & \textbf{DTS}$\downarrow$        \\ \cline{1-16}
\multicolumn{1}{l|}{\multirow{5}{*}{\rotatebox{90}{\textbf{Ablation}}}}                           & \textcolor{red}{\ding{56}}           & \textcolor{red}{\ding{56}}         & \textcolor{red}{\ding{56}}           & 45.74\%        & 0.340       & 1.577       & 34.02\%       & 0.278        & 2.371        & 30.87\%       & 0.251       & 3.052       & 36.48\%       & 0.287        & 2.391       \\
\multicolumn{1}{l|}{}                           & \textcolor{green}{\ding{52}}           & \textcolor{red}{\ding{56}}         & \textcolor{red}{\ding{56}}           & 55.09\%        & 0.404       & 1.464       & 39.41\%       & 0.326        & 2.396        & 37.86\%       & 0.316       & 2.957       & 43.53\%       & 0.316        & 2.346       \\
\multicolumn{1}{l|}{}                           & \textcolor{red}{\ding{56}}           & \textcolor{green}{\ding{52}}         & \textcolor{green}{\ding{52}}           & 52.13\%       & 0.367       & 1.605       & 42.31\%       & 0.319        & 2.203        & 44.40\%       & 0.324       & 3.048       & 46.25\%       & 0.336        & 2.232       \\
\multicolumn{1}{l|}{}                           & \textcolor{green}{\ding{52}}           & \textcolor{red}{\ding{56}}         & \textcolor{green}{\ding{52}}           & 54.41\%       & 0.368       & 1.546       & 42.21\%       & 0.272        & 2.096        & 42.10\%       & 0.292       & 3.032       & 45.49\%       & 0.307        & 2.204       \\
\multicolumn{1}{l|}{}                           & \textcolor{green}{\ding{52}}           & \textcolor{green}{\ding{52}}         & \textcolor{red}{\ding{56}}           & 62.23\%       & 0.422       & 1.406       & 51.48\%       & 0.352        & 1.847        & 47.15\%       & 0.328       & 2.794       & 53.15\%       & 0.365        & 2.190       \\ \hline
\multicolumn{1}{l|}{\textbf{Ours}}                           & \textcolor{green}{\ding{52}}           & \textcolor{green}{\ding{52}}         & \textcolor{green}{\ding{52}}             & \textbf{66.00\%}       & 0.458       & 1.289       & \textbf{59.50\%}       & 0.388        & \textbf{1.511}        & \textbf{54.75\%}       & \textbf{0.388}       & \textbf{2.684}       & \textbf{59.65\%}       & \textbf{0.385}        & \textbf{2.053} \\ \hline \hline
\end{tabular}
\end{adjustbox}

\label{tab:t1}
\vspace{-\baselineskip}
\vspace{-\baselineskip}
\vspace{-\baselineskip}
\end{table*}

\subsection{Multi-thinking Integration}

We use adaptive feature fusion for multi-thinking integration. Before feature fusion, we generate real-time perception features through visual inputs, which are crucial for the robot's obstacle avoidance capability. Here, we utilize a visual encoder to extract structural scene information.

To enable the balancing of exploratory and exploitative behaviors, we concatenate feature vectors $f_{cog}$, $f_u$, and $f_p$ into a fused feature $f_{{cat}_1}$. Subsequently, an attention layer with perceptrons calculates weights $w$ for each feature, which is then element-wise multiplied with $f_{{cat}_1}$ to produce $f_{{cat}_2}$, allowing the network to allocate attention adaptively. Afterward, $f_{{cat}_2}$ is fed into the navigation policy network to generate navigation action. The specific process is as follows:
\begin{equation}\label{e13}
f_{{cat}_2} = \textrm{MLP}(w\otimes f_{{cat}_1}),
\end{equation}
\begin{equation}\label{e14}
a = \textrm{Sample}(\textrm{Softmax}(\textrm{MLP}(f_{{cat}_2}))),
\end{equation}

\noindent where $f_{{cat}_1}=\textrm{Concat}(f_{cog},f_{u},f_{p})$, $w=\textrm{MLP}(f_{{cat}_1})$, and $\textrm{Sample}(\cdot)$ refers to the probability-based sampling.

%% file: sec/Experiment.tex
\section{Experiment}
\label{sec:Experiment}

\subsection{Implementation}

\noindent\textbf{Task Setup.} We conduct image-goal navigation tasks in iGibson~\cite{ref_59}. NUE is trained using imitation learning on the Gibson dataset~\cite{ref_60} with 21 scenes as training splits and 14 scenes as validation splits. These scenes are categorized into 3 difficulty levels based on the distance from the starting point to the target: 1) Easy: 1.5m - 3.0m; 2) Medium: 3.0m - 5.0m; 3) Hard: 5.0m - 10.0m. The robot has only access to the current RGBD observation and its current pose. The RGBD observation is obtained from a single RGBD camera with $180\times 240$ resolution and $90^{\circ}$ horizontal field of view. The maximum time step for each episode is set to 800. An event is considered successful when the robot reaches within a range of 0.8m of the target. Three evaluation metrics are used: success rate (SR), success weighted by path length (SPL), and distance to success (DTS).

\begin{figure*}[t] 
	\centering 
	\includegraphics[width=0.99\linewidth]{"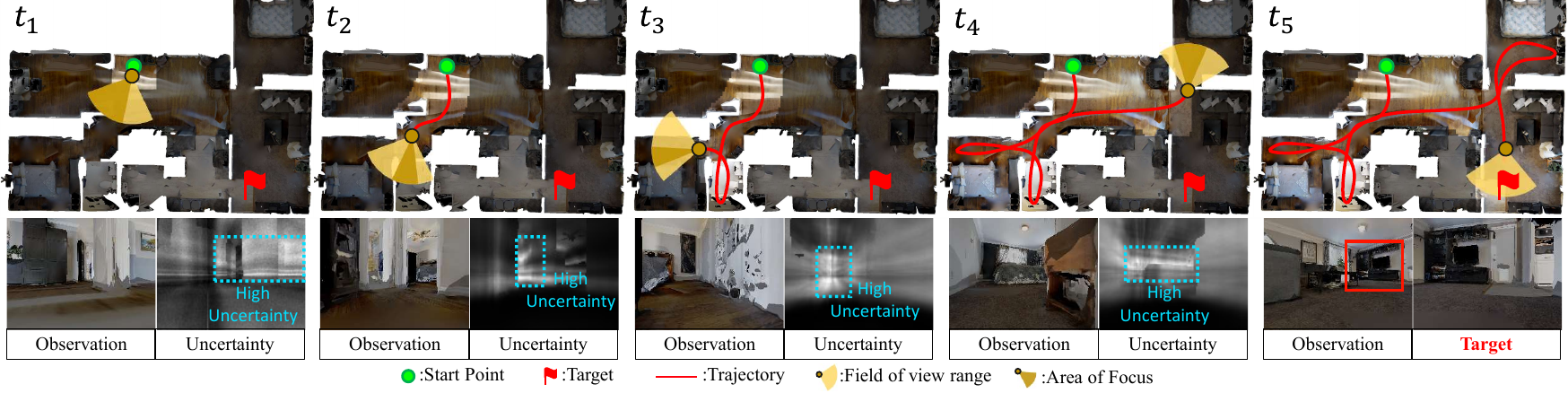"} 
	\caption{\textbf{Visualization examples of image-goal navigation.} The visualized results showcase the behavioral logic of our model. Our model successfully utilizes uncertainty to explore the scene in the early stages of navigation, while also effectively leveraging cognitive information for navigation upon sighting the target.}
    \label{fig_6}
\vspace{-\baselineskip}
\end{figure*}

\noindent\textbf{Baselines and Ablation Study.} We first compare the navigation efficiency of our model with the most basic models: Fully Reactive (\textbf{FR})~\cite{ref_61} and \textbf{Nav-A3C}~\cite{ref_16}. In addition, to validate the effectiveness of our cognitive structure, we include \textbf{VGM}~\cite{ref_27} and Multi-Storage Memory (\textbf{MSM})~\cite{ref_62} as baselines. We also compare our method with recent state-of-the-art image-goal navigation approaches. \textbf{NRNS}~\cite{ref_64} constructs a topological map by predicting the distance between the target image and node images. \textbf{SLING}~\cite{ref_66} enhances the robot's navigation capability in the ``last mile'' by predicting the relative pose of the target based on keypoint matching. We combine SLING with the reinforcement learning-based \textbf{DDPPO}~\cite{ref_67} and \textbf{OVRL}~\cite{ref_68} as baselines. In ablation study, we conduct ablation on 3 parts: 1) $f_u$: we remove $f_u$, and only fuse $f_{cog}$ and $f_p$ to generate actions; 2) \textit{auxiliary task} (AT): we delete the auxiliary task for target direction prediction from the full model; 3) CBAM: we remove the CBAM layer and rely solely on the residual network and MLPs for cognitive feature extraction.

\subsection{Image-goal navigation Results}
Table~\ref{tab:t1} presents the average SR, SPL, and DTS for each method. Compared to FR and Nav-A3C, our model achieves significantly enhanced navigation performance, highlighting the significance of cognitive structure. Moreover, our model outperforms baselines with internal scene representation (MSM, VGM, NRNS), showcasing the effectiveness of NeRF as a scene representation method and the value of our information extraction approach. Furthermore, compared to reinforcement learning methods enhanced by last-mile navigation (SLING+DDPPO, SLING+OVRL), our model still exhibits superior performance in hard scenarios, which are more exploration-dependent. This underscores the benefits of using uncertainty to enhance exploration behaviors for robot navigation in complex environments. 

The ablation study is also shown in Table~\ref{tab:t1}. The model only with the addition of $f_u$ demonstrated significantly better performance compared to the fully ablated model. Meanwhile, The removal of $f_u$ led to a significant decrease in the navigation performance of our model, demonstrating that our approach of enhancing exploration using uncertainty improves the robot's ability to establish cognitive understanding. Similarly, the ablation of AT also resulted in a decrease in navigation performance, indicating that our auxiliary task effectively improves the robot's reasoning capability for determining the target's position. Furthermore, after ablating the CBAM layer, we can observe a decrease in the navigation performance of our model, indicating that CBAM attention contributes to improving feature extraction effectiveness.

\subsection{Interpretability Experiments}
\subsubsection{Visualization of Typical Navigation Behaviors}

In Fig.~\ref{fig_6}, we visualize the robot’s trajectory in a typical testing episode. At the initial stage of this task ($t_1$), the cognitive structure has limited environmental memory, resulting in low signal-to-noise ratio outputs. At this phase, uncertainty information provides additional assistance to the robot's decision-making. As the robot observes a new room ($t_2$, $t_3$, $t_4$), higher uncertainty is observed in unexplored areas, guiding the robot to explore new regions and avoiding redundant paths. Finally, when the robot encounters an area similar to the target image ($t_5$), it decisively chooses to move toward it. This experiment confirms that our pipeline effectively achieves the transition from exploration to exploitation.

\subsubsection{Results of Auxiliary Task}

\begin{figure}[t] 
	\centering 
	\includegraphics[width=1\linewidth]{"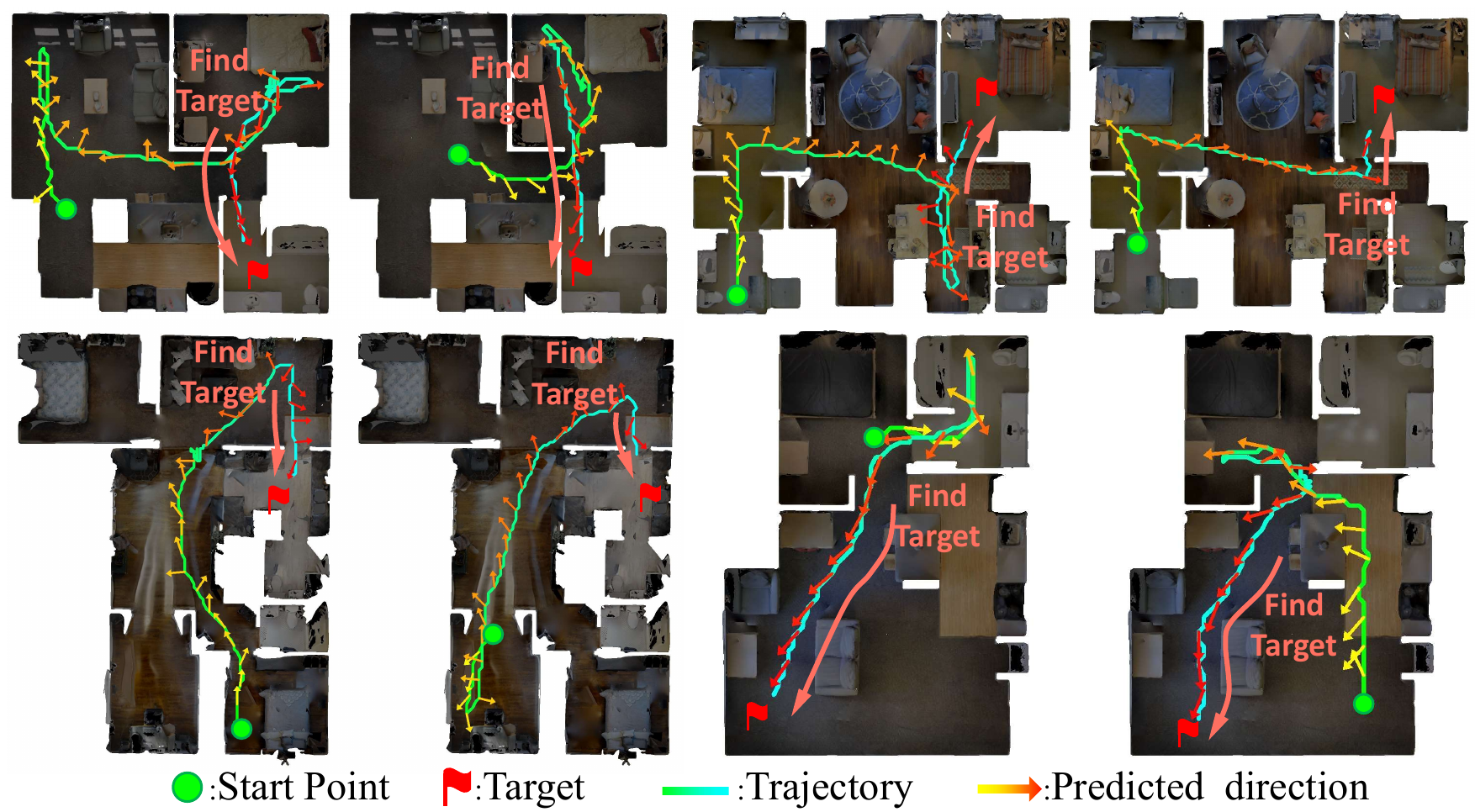"} 
	\caption{\textbf{Predicted results of the auxiliary task.} The change in arrow color from yellow to red indicates the progress of navigation. During navigation tasks, auxiliary task accuracy steadily enhances, especially after observing the target.}
    \label{fig_7}
\vspace{-\baselineskip}
\vspace{-\baselineskip}
\vspace{-\baselineskip}
\end{figure}

To assess the auxiliary task's efficacy, we select 8 navigation tasks across 4 scenes and visualize the predicted target direction in Fig.~\ref{fig_7}. Initially, with limited robot exploration range and lack of target-related cues, the cognitive structure lacks target memory, resulting in subpar prediction performance. A noticeable mismatch between predicted and actual robot actions suggests exploratory thinking guiding decision-making. However, as the robot's exploration expands, auxiliary task accuracy notably improves. At this stage, exploitative thinking prevails, effectively steering the robot towards the target. The visualization confirms the auxiliary task's role in enhancing the robot's reasoning for the target position.

\subsubsection{Interpretation of Uncertainty Extraction}

\begin{figure}[t] 
	\centering 
	\includegraphics[width=0.95\linewidth]{"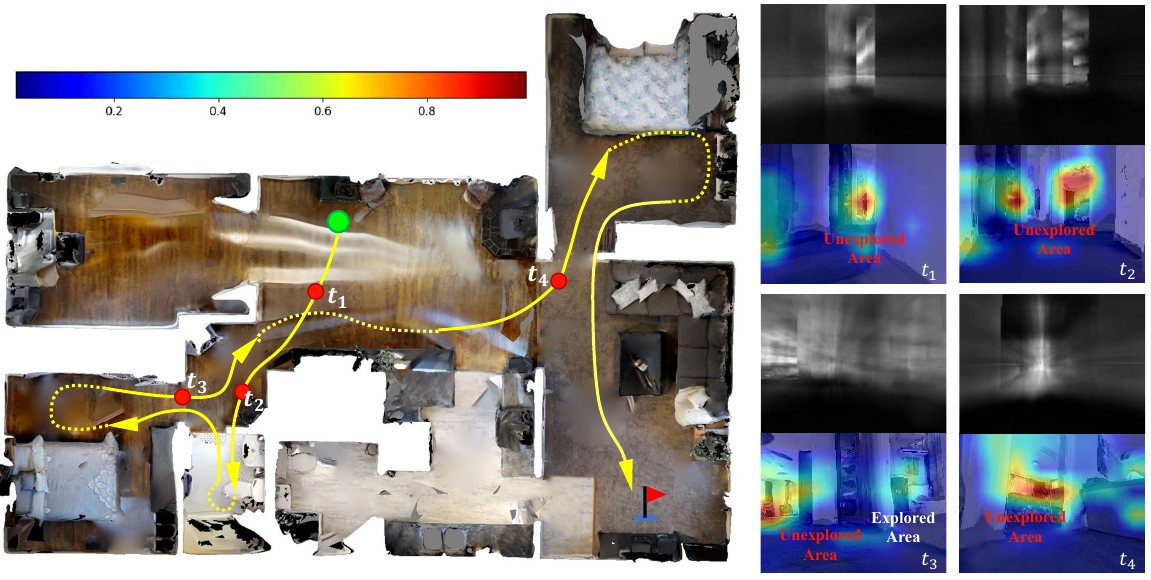"} 
	\caption{\textbf{Interpretation results of uncertainty extraction.} The color in the image signifies the attention level of our uncertainty extraction structure on various regions. The top image in each pair shows the uncertainty map generated by NeRF, while the bottom image displays the gradient heatmap used for policy generation.}
    \label{fig_8}
\vspace{-\baselineskip}
\end{figure}

Fig.~\ref{fig_8} showcases the gradient visualization of our uncertainty extraction structure using Grad-CAM. The color gradient from blue to red represents the ascending order of the model's attention intensity. It can be observed that the model can effectively allocate high attention to regions with high uncertainty, which indicates that our uncertainty extraction structure can focus on unexplored regions through the uncertainty map, thus guiding the robot to explore these areas.

%% file: sec/Conclusion.tex
\section{Conclusion}
\label{sec:Conclusion}

We introduce an end-to-end visuomotor navigation framework, NUE, which applies the powerful scene representation capability of NeRF to the field of image-goal navigation, and specifically enhances the robot's exploratory behavior through uncertainty estimation and extraction. This innovative approach overcomes the problem of insufficient focus on exploratory behavior in traditional methods, enabling the robot to rapidly establish environmental cognition to provide more information for the subsequent exploitation phase. Our experiments provide compelling evidence that our framework has achieved active exploration behavior for the robot, thereby improving the efficiency of navigation.